\documentclass[letterpaper]{article}
\usepackage{uai2020}
\usepackage[margin=1in]{geometry}

\usepackage{times}
\usepackage{graphicx}
\usepackage{url}
\usepackage{algorithm2e}
\usepackage{algorithmic}

\newcommand{\Policy}{\pi_{\theta}}

\newcommand{\E}{\mathbb{E}}

\usepackage{xcolor}
\usepackage{hyperref}
\hypersetup{
    colorlinks=true,
    linkcolor=blue,
    citecolor=blue,
    filecolor=magenta,      
    urlcolor=cyan,
    hyperindex=True,
 urlcolor=blue}

\newcommand{\hide}[1]{}

\usepackage{amsmath,amsfonts,bm}

\def\eqref#1{equation~\ref{#1}}

\def\1{\bm{1}}

\def\va{{\bm{a}}}

\def\vc{{\bm{c}}}

\def\vh{{\bm{h}}}

\def\vs{{\bm{s}}}

\def\vz{{\bm{z}}}

\def\mB{{\bm{B}}}
\def\mC{{\bm{C}}}

\DeclareMathAlphabet{\mathsfit}{\encodingdefault}{\sfdefault}{m}{sl}
\SetMathAlphabet{\mathsfit}{bold}{\encodingdefault}{\sfdefault}{bx}{n}

\def\gA{{\mathcal{A}}}
\def\gB{{\mathcal{B}}}

\def\gS{{\mathcal{S}}}
\def\gT{{\mathcal{T}}}

\newcommand{\Q}{Q}

\usepackage{xspace}
\newcommand{\methodname}{\textsc{Ocean}\xspace}
\newcommand{\pearl}{\textsc{Pearl}\xspace}

\def\zglobal{{\vz^{{\sc Global}}}}
\def\zlocal{{\vz^{{\sc Local}}}}
\def\zlocalzero{{\vz^{{\sc Local}}_0}}
\def\zlocaltplus{{\vz^{{\sc Local}}_{t+1}}}
\def\zlocaltr{{\vz^{{\sc Local}}_{tr}}}
\def\zlocalt{{\vz^{{\sc Local}}_t}}

\def\localenc{{q_\phi^L}}
\def\globalenc{{q_\phi^G}}
\def\cenc{{q_\phi^\text{enc}}}

\def\tran{{q_\phi^\text{tran}}}
\def\zprior{{q_\phi^\text{prior}}}

\title{OCEAN: Online Task Inference for Compositional Tasks \\with Context Adaptation}

\author{ {\bf Hongyu Ren} \\
Stanford University\\
\And
{\bf Yuke Zhu}  \\
The University of Texas at Austin, Nvidia \\
\And
{\bf Jure Leskovec}   \\
Stanford University    \\
\AND
{\bf Anima Anandkumar} \\
California Institute of Technology, Nvidia \\
\And
{\bf Animesh Garg} \\
University of Toronto, Vector Institute, Nvidia  \\
}

\begin{document}

\maketitle

\begin{abstract}
Real-world tasks often exhibit a compositional structure that contains a sequence of simpler sub-tasks. For instance, opening a door requires reaching, grasping, rotating, and pulling the door knob. Such compositional tasks require an agent to reason about the sub-task \emph{at hand} while orchestrating \emph{global} behavior accordingly. This can be cast as an \emph{online task inference} problem, where the current task identity, represented by a context variable, is estimated from the agent's past experiences with probabilistic inference.
Previous approaches have employed simple latent distributions, e.g., Gaussian, to model a single context for the entire task. However, this formulation lacks the expressiveness to capture the composition and transition of the sub-tasks. 
We propose a variational inference framework \methodname to perform online task inference for compositional tasks. \methodname models global and local context variables in a joint latent space, where the global variables represent a mixture of sub-tasks required for the task, while the local variables capture the transitions between the sub-tasks. 
Our framework supports flexible latent distributions based on prior knowledge of the task structure and can be trained in an unsupervised manner. 
Experimental results show that \methodname provides more effective task inference with sequential context adaptation and thus leads to a performance boost on complex, multi-stage tasks.
\end{abstract}

\section{INTRODUCTION}
Meta-reinforcement learning (meta-RL) algorithms  aim to train a versatile agent that quickly adapts to unseen new tasks after trained on a domain of related tasks \cite{maml,emaml,promp,rl2,pearl}. Meta-RL has shown promise in simple domains, where it requires much less data than training an RL agent from scratch for each task of interest. A key assumption that makes meta-RL desirable is that there exists rich latent structure and relationships among the tasks, implying that solving one task may reuse skills from others. For instance, the agent learning to walk backward may benefit from mastering how to run forward. However, leveraging this structure in task inference and adaptation \cite{taskinference} makes meta-RL much more challenging than RL, albeit offers an opportunity for improvement in sample efficiency of multi-task learning.

\begin{figure*}[t]
\label{fig:method}
\centering
    \includegraphics[width=\textwidth]{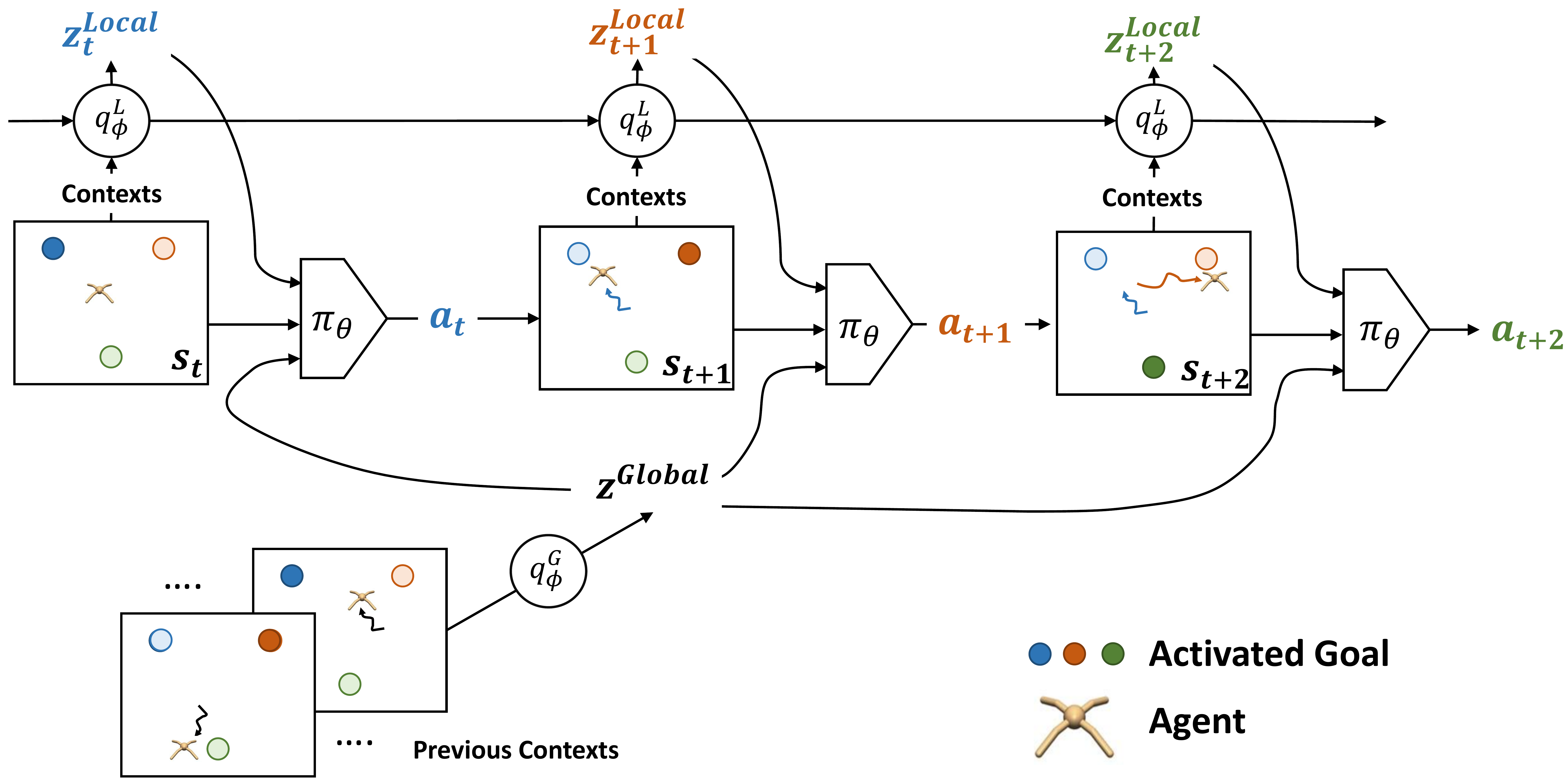}
    \caption{Our framework \methodname performs task inference based on the contexts. \methodname consists of two modules: global context encoder $\globalenc$ and local context encoder $\localenc$. $\globalenc$ leverages contexts from previous episodes to capture the global structure of the task, for example, goal locations, rewards. $\localenc$ (with a recurrent architecture) uses the contexts from previous steps within an episode to do online sub-task inference, it aims to capture the current sub-task, in this case, currently which goal has been activated. The agent takes actions $\va_t$ based on the current state $\vs_t$, the current local latent context variable $\zlocalt$ and the global context variable $\zglobal$.}
\end{figure*}

Recent context-based meta-RL models \cite{pearl,iclrpearl} capture the current task with additional latent task variables. Specifically, the agent first stochastically explores the environment and keeps record of (potentially unordered) transition tuples, which are referred to as \emph{contexts}. A trained context encoder uses the contexts to infer the latent task variables, which serve as the agent's belief over the current task. Given the latent task variable, the agent makes sequential decisions for the current task. These context-based meta-RL models provide an elegant framework that disentangles probabilistic task inference from decision making, and can be integrated with multiple off-policy learning algorithms \cite{sac,haarnoja2018soft,mnih2015human}, demonstrating improvement in both sample efficiency and asymptotic performance. 

However, prior art fails in case of complex task structures in most real-world scenarios. Often real tasks require finishing a sequence of sub-tasks and the current sub-task may only be revealed to the agent after the previous sub-task is finished. For example, door opening requires one to reach, grasp, rotate and pull or push the door knob. Not only do these sub-tasks need to be carried out in an appropriate order, but the each subsequent sub-task is also conditioned on the success previous one.  
This compositional structure is of important value and needs to be accounted for during task inference. However, current context-based meta-RL models do not leverage this sophisticated multi-stage structure and the dependencies between tasks. The reason is modelling the whole episode of a task with a fixed isotropic Gaussian context variables. As shown in Fig. \ref{fig:method}, one task may consist of multiple stages where each stage may have a different goal, a single static latent context variable fails to model the sequential structure. Although posterior estimation may offer slight improvement through recurrence as more contexts accumulate, the following issues remain: (1) latent context variables that are held constant across an episode are not suitable to model the sequence of sub-tasks executed in an order; (2) isotropic Gaussian random variables are not flexible enough to model mixtures of tasks. 

This paper proposes \methodname, an Online ContExt AdaptatioN framework that addresses the aforementioned issues \footnote{The implementation can be found in \url{https://github.com/pairlab/ocean}.}. Our framework is composed of two parts that account for the local sub-task update and the global sub-task mixture respectively. We introduce a local context encoder with a recurrent architecture, which performs sub-task inference on-the-fly based on the history contexts within an episode as shown in top of Fig. \ref{fig:method}. Vitally, the adaptive local context variables instruct the agent to make smooth transitions from one sub-task to another within a single episode. In order to model different sub-task structure, we also introduce a global context encoder facilitates a flexible latent space with rich prior distributions beyond Gaussian, categorical, Dirichlet and logistic normal distributions based on the domain knowledge of the task structure as shown in bottom of Fig. \ref{fig:method}. With a joint latent space comprised of the local and global context variables, \methodname gives rise to \emph{a principled task inference framework for context-based meta-reinforcement learning} that is able to model complex, real-world tasks. We observe that previous context-based meta-RL models \cite{pearl,iclrpearl} can be viewed as special cases of our general framework where global latent space is Gaussian and the local context variables are frozen.

We integrate \methodname with soft actor-critic \cite{sac}, an off-policy RL algorithm with high sample efficiency. The context encoders and the RL agent can be trained jointly to optimize for maximizing expected return across all the tasks in meta-training. Note that our framework does not require any labels or supervised signals of either sub-task mixture or sub-task transition when training both the global and local encoder. In meta-test phase, we sample the global context variables for a given new task, and stochastically explore the task with both the global context variables and the local context variables, which \methodname estimate and update in an online fashion. Task inference becomes more and more accurate with more exploration, thus adapting to the new task in meta-test.
We show through a 2D locomotion experiment that designing an appropriate global latent space matters for complex task structures. We further demonstrate that online context adaptation can greatly improve the performance in several continuous control tasks with sequential sub-task structure.

\section{PRELIMINARIES}

\subsection{META-REINFORCEMENT LEARNING}
We are interested in meta-reinforcement learning (meta-RL), where we are given a distribution of tasks $p(\gT)$. Each sample from $p(\gT)$ is a Markov Decision Process (MDP) $\langle\gS, \gA, P, r, \rho_0, \gamma\rangle$, representing the state space, action space, transition probability distribution, reward function, the distribution of initial state and discount factor respectively. We assume that all tasks from $p(\gT)$ share the same known state and action space, but may differ in transition probability, reward function and initial state distribution, which are unknown but we can sample from them. We refer to the contexts $\mC$ of a given task $\gT$ as a collection of transition tuples sampled from $\gT$. Each transition tuple $(\vs_i, \va_i, r_i, \vs'_i)$ consists of the state $\vs$, action $\va$, reward $r$, next state $\vs'$ at a certain timestep. Meta-RL models aim to train a flexible agent $\pi_\theta(\va|\vs)$ (parameterized by $\theta$) on a given set of training tasks sampled from $p(\gT)$ and the goal is that the agent can be adapted quickly to a 

\subsection{CONTEXT-BASED TASK INFERENCE}
In order to make fast adaptation towards a given task, the key is to perform accurate task inference, which can be explicitly captured by latent task variables $p(\vz|\gT)$.
The RL agent $\pi_\theta(\va|\vs, \vz)$ then takes actions based on the current observation (or state) $\vs$ and the latent task variables $\vz \sim p(\vz|\gT)$ in order to make decisions in task $\gT$. The latent task variables capture the uncertainty over the task and are crucial to achieve quick adaptation and high performance in meta-RL. However, the true posterior is unknown since we have no knowledge of the transition probability, reward function and initial state distribution of the task $\gT$. Context-based meta-RL models approximate $p(\vz|\gT)$ by first collecting a set of contexts $\mC$ from the task $\gT$ and calculating the posterior of latent \emph{context} variable $p(\vz|\mC)$, where essentially the contexts can be seen as the representative samples from the task $\gT$. Although $p(\vz|\mC)$ is still intractable, we can train an additional context encoder $q_\phi(\vz|\mC)$ parameterized by $\phi$ to estimate the true posterior based on amortized variational inference. The corresponding evidence lower bound can be derived as 
\begin{align}\label{eq:pearl}
\mathbb{E}_{\gT} [\mathbb{E}_{\vz \sim q_\phi}[R(\gT, \vz)
- \beta D_{\text{KL}}(q_\phi(\vz | \mC) || p(\vz))]],
\end{align}
where $p(\vz)$ is the prior distribution and captures our prior knowledge of the task distribution, $R(\gT, \vz)$ is implemented to recover the state-action value functions \cite{pearl} or the transition functions \cite{iclrpearl}, $\beta$ is a trade-off hyperparameter that regularizes the capacity. One benefit of context-based meta-RL models is that it disentangles task inference from decision making and hence large amount of off-policy data can be used for policy update, which greatly improves the sample efficiency \cite{pearl}. 

\subsection{SOFT ACTOR-CRITIC}\label{sec:sac}
In this paper, we use the soft actor-critic algorithm (SAC) \cite{sac,haarnoja2018soft} to perform policy update in order to achieve better sample efficiency. Based on the maximum entropy RL framework \cite{ziebart2008maximum}, SAC is an off-policy actor-critic algorithm that aims to maximize both the expected reward and casual entropy, explicitly regularizing the policy. Specifically, SAC aims to optimize the following objectives for the agent $\pi_\theta$, the Q-function approximator $Q_\theta$ and the value function approximator $V_\theta$ using samples stored in a replay buffer $\gB$:
\begin{equation}\label{eq:critic}
\mathcal{L}_{critic} = \E_{(\vs, \va, \vs', r) \sim \gB} [Q_\theta(\vs, \va) - (r + V_\theta(\vs'))]^2, \\
\end{equation}
\begin{equation}\label{eq:actor}
\mathcal{L}_{actor} \!=\! \E_{\vs \sim \gB} \!\left[\! D_{\text{KL}}\!\left(\!\Policy(\cdot | \vs) \! \left\| \frac{\text{exp}\left(Q_\theta(\vs, \cdot)\right)}{\mathcal{Z}_\theta(\vs)} \! \right. \right) \! \right].
\end{equation}
\section{\methodname: TASK INFERENCE WITH LATENT CONTEXT VARIABLES}

Latent context variables are crucial in context-based meta RL models since they represent the belief over the current tasks. Thus, these context variables should be carefully tailored during task inference to take into account the complex compositional structure of real-world tasks in order to achieve the most accurate context variables.  
In Sec. \ref{sec:online}, we introduce \emph{local} latent context variables and perform online context adaptation and sub-task inference based on the contexts at past steps within an episode. We implement a local context encoder with a recurrent neural network for sequential probabilistic inference. In Sec. \ref{sec:global}, we use \emph{global} context variables to capture the global information of a task and further propose \methodname, an Online ContExt AdaptatioN framework for meta-RL with joint latent space that consists of global and local context variables. 
In Sec. \ref{sec:flexible}, we demonstrate that our framework allows flexible latent distributions to suit different prior knowledge of the task/sub-task structure.
Finally, in Sec. \ref{sec:training}, we integrate our task inference framework with SAC for efficient policy update, and perform end-to-end training for the global and local context encoders as well as the agent.

\subsection{ONLINE CONTEXT ADAPTATION}\label{sec:online}
Since real-world tasks often require an agent to finish a sequence of sub-tasks, a single static latent context variable is not sufficient to inform the agent of the transitions in the sequence. 
To capture the ongoing sub-task and the transitions between them, we use local latent context variables $\zlocal$ that are estimated online throughout the episode. To perform online probabilistic inference, we train an additional local context encoder $\localenc$ parameterized by $\pi$ that takes as input the contexts at previous steps and updates the posterior of the local context variable for future steps. Since the local latent contexts at different steps are not independent from each other, we design a recurrent architecture for $\localenc$. 
To better model the variability of dependencies between the local context variables across different timesteps, we adopt variational recurrent neural network \cite{vrnn}, where the hidden state is conditioned on stochastic samples from the previous posterior. Specifically, the local context encoder $\localenc$ consists of three modules: $\cenc$, $\tran$ and $\zprior$, which represent the inference function, the transition function, the conditional prior respectively. Given the context $\vc_{t}$ at timestep $t$, the latent context variable $\zlocaltplus$ at timestep $t+1$ is sampled from the posterior calculated as follows:
\begin{align}\label{eq:enc}
    \zlocaltplus \sim \cenc(\vz|\vc_{t}, \vh_{t}),
\end{align}
where $\vh_{t}$ denotes the hidden state at timestep $t$. The hidden state is updated according to the following recurrence:
\begin{align}\label{eq:tran}
    \vh_{t} = \tran(\vc_{t-1}, \zlocalt, \vh_{t-1}).
\end{align}
Then the agent takes actions by sampling from $\pi_\theta(\va|\vs_t,\zlocalt)$, which is conditioned on the observation as well as the updated local context variable at timestep $t$. Note that we use zero-valued vectors as initialization for $\vh_0$, and we can directly sample $\zlocalzero$ from a predefined uninformative prior, such as isotropic Gaussian or uniform  distribution. 

Since we aim to capture the dependency between the current and past timesteps, the corresponding prior distribution of $\zlocalt$ is conditioned on the previous hidden state $p(\zlocalt)=\zprior(\vh_{t-1})$ rather than a given fixed uninformative prior, which neglects the temporal structure of posterior at different steps. 

Finally the loss of the local context encoder $\localenc$ is defined by replacing the KL loss term in Eq. \ref{eq:pearl} as follows:
\begin{align}
    D_\text{KL}^{Local} = \sum_t{D_\text{KL}(\cenc(\vz|\vc_t,\vh_t)||\zprior(\vh_t))},\label{eq:localkl}
\end{align}
which takes the sum of the KL loss at all timesteps. and is optimized in meta-training.
Note that \methodname does not assume prior knowledge of either the labels of the specific transition steps between each sub-task or the labels of the sub-tasks. Using the variational inference framework, our model is able to discover sub-task structure in an unsupervised manner.
At meta-test time, we fix the local context encoder $\localenc$ and the agent first samples from the prior distribution and then can execute the policy, collect the context and infer the posterior of the local context variable in a recurrent manner. Given a previously unseen test task, the agent explores at the first few steps and then with more steps collected, quickly converges to the optimal policy as the tasks inference becomes more and more accurate, thus efficiently adapted to the test task at hand. 

\textbf{Computation Efficiency.}
Note that the local context encoder requires stochastic sampling at each step, which is a sequential process. Given a limited computation budget, it can be costly. 
Although real-world tasks often contain sequence of sub-tasks, we do not need to update the posterior at each step. We assume that the sub-task in timestep $t$ is very similar to the sub-task in timestep $t+1$, so one local context variable that represents the current sub-task at timestep $t$ is also likely to be accurate enough for decision-making at timestep $t+1$.
To reduce the computation overhead, one strategy is to perform posterior estimation on a subset of the timesteps $\{tr, 2tr, \dots\}$ with certain temporal resolution $tr$. Specifically for Eq. \ref{eq:enc}, $\zlocaltr$ is sampled from $\cenc(\vz|\vc_{0:tr-1},\vh_0)$, where $\vc_{0:tr-1}$ is the concatenated contexts from timestep 0 to $tr-1$. 
Then the cost of each posterior sampling can be amortized over $tr$ steps of decision making.

\subsection{JOINT LATENT SPACE FOR TASK INFERENCE}\label{sec:global}
As introduced in Sec. \ref{sec:online}, given a task, the agent first samples from uninformative prior and explores the environment at the beginning of the episode in order to infer the local sub-task with $\localenc$, however, these exploration steps may be sub-optimal since the agent has no knowledge of the task and explores with context variables drawn from uninformative prior. One limitation of the local context variables is that they fail to make use of past contexts from different trajectories. In this section, we introduce global context variables which leverages the contexts from past trajectories to give the agent a global overview of the task. 
If we have access to a pool of past contexts $\mC$ of size $n$ collected from the same task (but not necessarily from the same episode), we leverage this task information to infer our belief over the global task. Following \pearl \cite{pearl}, we introduce a global context encoder $\globalenc(\vz|\vc_i)$ that infers the posterior based on each single past context $\vc_i \in \mC$. The posterior of the global context is calculated as a function of each independent posterior $\globalenc(\vz|\mC) = f(\globalenc(\vz|\vc_1), \dots, \globalenc(\vz|\vc_n))$ (detailed in Sec. \ref{sec:flexible}). With larger set of the past contexts $\mC$, we can achieve more accurate global task estimation.

Combined with the local context variables, we introduce \methodname with a joint latent space for online task inference in meta-RL. The joint latent space consists of local and global context variables where the local context variables reason about the sub-task and are updated online, while the global context variables captures the big picture of the task. The global and local context encoder can be trained jointly with the agent with detailed description in Sec. \ref{sec:training}. For simplicity, we assume the local and global context variables are independent, thus the KL term in the objective in Eq. \ref{eq:pearl} can be decomposed into sum of two separate terms:
\begin{align}
    D_\text{KL}(\globalenc(\vz|\mC)||p(\vz)) + D_\text{KL}^{Local},\label{eq:allkl}
\end{align}
where $p(\vz)$ is the prior distribution for global context variables and $D_\text{KL}^{Local}$ is defined in Eq. \ref{eq:localkl}.

\begin{table}[!t]
\resizebox{\columnwidth}{!}{
\begin{tabular}{l}
\hline 
\textbf{Algorithm 1: Meta-Training in OCEAN} \\
\hline
 \textbf{Input:} $\gT_{1\dots T}$: Training tasks sampled from $p(\gT)$. \\
 \textbf{Initialize:} $\gB^i$: Replay buffers for each task; \\
 $\theta_\pi,\theta_Q,\theta_V, \phi$: parameters in \methodname; $\alpha_1, \alpha_2, \alpha_3$: learning rate; \\
 $K$: number of trajectories collected in each iteration; \\
 $B$: number of trajectories sampled in each iteration. \\
01. \textbf{While} not done \textbf{do} \\
02.\quad\vline height 2ex\; \textbf{For} $i = 1, \dots, T$ \textbf{do} \\
03.\quad\vline height 2ex\quad\vline height 2ex\; TrajCollect($\gT_i$, $\gB^i$, $K$) \\ 
04.\quad\vline height 2ex\quad\vline height 2ex\; Contexts $\mC \sim \mathcal{S}_c(\mathcal{B}^i)$ and trajectories $\mB \sim \mathcal{B}^{i}$ \\ 
05.\quad\vline height 2ex\quad\vline height 2ex\; $\zglobal \sim \globalenc(\vz | \mC)$ \\ 
06.\quad\vline height 2ex\quad\vline height 2ex\; Initialize $\zlocal = \{\}$ \\ 
07.\quad\vline height 2ex\quad\vline height 2ex\; $\mathcal{L}^i_{KL} = \beta D_{\text{KL}}(q(\vz | \mC) || p(\vz))$ \\ 
08.\quad\vline height 2ex\quad\vline height 2ex\; \textbf{For} $b = 1, \dots, B$ \textbf{do} \\
09.\quad\vline height 2ex\quad\vline height 2ex\quad\vline height 2ex\; \textbf{For} $t = 1,2, \dots$ \textbf{do} \\
10.\quad\vline height 2ex\quad\vline height 2ex\quad\vline height 2ex\quad\vline height 2ex\; $\zlocalt \sim \localenc(\vz | \vc_{t-1}, \vh_{t-1})$ and add to $\zlocal$ \\
11.\quad\vline height 2ex\quad\vline height 2ex\quad\vline height 2ex\quad\vline height 2ex\; Update $\vc_t = (\vs_t, \va_t, \vs'_t, r_t)$ \\
12.\quad\vline height 2ex\quad\vline height 2ex\quad\vline height 2ex depth 0.5 pt \line(1,0){10} \vline height 2ex depth 0.5 pt \line(1,0){2.5}\; $\mathcal{L}^i_{KL} += \beta D_{\text{KL}}(\localenc( \vz | \vc_{t-1}, \vh_{t-1}) || p(\zlocalt))$ \\
13.\quad\vline height 2ex\quad\vline height 2ex\; $\mathcal{L}^i_{actor} = \mathcal{L}_{actor}(\mB, \zglobal, \zlocal)$ \\
14.\quad\vline height 2ex\quad\vline height 2ex depth 0.5 pt \line(1,0){2.5}\; $\mathcal{L}^i_{critic} = \mathcal{L}_{critic}(\mB, \zglobal, \zlocal)$ \\
15.\quad\vline height 2ex\; $\phi \gets \phi - \alpha_1 \nabla_\phi \sum_i \left(\mathcal{L}^i_{critic} + \mathcal{L}^i_{KL}\right)$ \\
16.\quad\vline height 2ex\; $\theta_{\pi} \gets \theta_{\pi} - \alpha_2  \nabla_{\theta_\pi} \sum_i \mathcal{L}^i_{actor}$ \\
17.\quad\vline height 2ex\; $\theta_{Q} \gets \theta_{Q} - \alpha_3  \nabla_{\theta_Q} \sum_i \mathcal{L}^i_{critic}$ \\
18.\quad\vline height 2ex depth 0.5 pt \line(1,0){2.5}\; $\theta_{V} \gets \theta_{V} - \alpha_3  \nabla_{\theta_V} \sum_i \mathcal{L}^i_{critic}$ \\
\hline 
\end{tabular}
}
\vspace{-8pt}
\end{table}

\begin{table}[!t]
\resizebox{\columnwidth}{!}{
\begin{tabular}{l}
\hline 
\textbf{Algorithm 2: TrajCollect} \\
\hline
 \textbf{Input:} $\gT$: Task; $\gB$: Replay Buffer; $K$: Number of trajectories. \\
 \textbf{Initialize:} $\mC$: Context set. \\
01. \textbf{For} $k = 1, \dots, K$ \textbf{do} \\
02.\quad\vline height 2ex\; $\zglobal \sim \globalenc(\vz | \mC)$ \\
03.\quad\vline height 2ex\; \textbf{For} $t = 1,2, \dots$ \textbf{do} \\
04.\quad\vline height 2ex\quad\vline height 2ex\; $\zlocalt \sim \localenc(\vz | \vc_{t-1}, \vh_{t-1})$ \\ 
05.\quad\vline height 2ex\quad\vline height 2ex\; Roll out policy $\pi_{\theta}(\va | \vs, \zglobal, \zlocalt)$  \\ 
06.\quad\vline height 2ex\quad\vline height 2ex\; Update $\vc_t = (\vs_t, \va_t, \vs'_t, r_t)$ \\ 
07.\quad\vline height 2ex depth 0.5 pt \line(1,0){10} \vline height 2ex depth 0.5 pt \line(1,0){2.5}\; Accumulate context $\mC = \mC \cup \{\vc_t\}$ \\ 
08. Add $\mC$ to $\gB$ \\ 
\hline 
\end{tabular}
}
\vspace{-5pt}
\end{table}

\subsection{FLEXIBLE PARAMETERIZATION OF THE LATENT SPACE}\label{sec:flexible}
A suitable latent space is critical in accurate task inference. Based on the prior knowledge of the task, \methodname supports flexible parameterization of the latent space for both the global and the local context variables. Besides the Gaussian prior used in \pearl \cite{pearl}, we also design latent space with categorical distribution to model tasks controlled by discrete factors, Dirichlet distribution and logistic normal (logit-normal) distribution to model multi-modal or proportional tasks. While it is straight forward for local context encoder to infer the posterior for the next step, estimating the global posterior is non-trivial since we need to consider all the past contexts from the replay buffer and design a suitable function $f$ as introduced in Sec. \ref{sec:global}.

When estimating the posterior of global latent context variables, we model each context as an independent factor in order to capture the minimal sufficient information \cite{pearl}. Specifically, assume we have $n$ contexts, we first estimate the posterior $\globalenc(\vz|\vc_i)$ for each single context $\vc_i$ using the global context encoder $\globalenc$. Then we model the global latent context variable $\globalenc(\vz|\mC)$ as the weighted product of the independent posteriors: $\globalenc(\vz|\mC)\propto \prod_{i=1}^n \globalenc(\vz|\vc_i)^\frac{1}{n}$. Our framework allows the global latent space with Gaussian distribution, categorical distribution, Dirichlet distribution and logit-normal distribution. For all four distributions, the operation of weighted product of the probability density functions (PDFs) is closed, and thus the we can easily calculate the global posterior as shown below.

\textbf{Gaussian / Logit-normal Distribution.} Assume the posterior of context $\vc_i$ has parameters $\mu_i$ and $\sigma_i^2$, then the global posterior has parameters $\mu=\frac{1}{\sum_i \frac{1}{\sigma_i^2}}\sum_i \frac{\mu_i^2}{\sigma_i^2}$ and $\sigma^2=\frac{n}{\sum_i \frac{1}{\sigma_i^2}}$. 

\textbf{Categorical Distribution.} Assume the posterior of context $\vc_i$ has parameters $(p_{i1},\dots, p_{iK})$, the global posterior has parameters $(\frac{\sqrt[n]{\prod_i p_{i1}}}{Z}, \dots, \frac{\sqrt[n]{\prod_i p_{iK}}}{Z})$, where $Z=\sum_j \sqrt[n]{\prod_i p_{ij}}$.

\textbf{Dirichlet Distribution.} Assume the posterior of context $\vc_i$ has parameters $(\alpha_{i1}, \dots, \alpha_{iK})$, the global posterior has parameters $(\frac{\sum_i \alpha_{i1}}{n}, \dots, \frac{\sum_i \alpha_{iK}}{n})$.

After we achieve the posterior of global and local contexts, we can use the reparameterization trick to sample from these distributions in order to optimize the Eq. \ref{eq:allkl}. Specifically, we use several reparameterization tricks for Gaussian \cite{vae}, categorical distribution \cite{gumbel,concrete} and Dirichlet distribution \cite{implicit}.  
Note that the latent space can be composite and may consist of random variables from different distributions mentioned above based on the prior knowledge of the tasks.

\subsection{TRAINING WITH OFF-POLICY UPDATE}\label{sec:training}
Sample efficiency is one of the most critical issues in both RL and meta-RL. Following \pearl \cite{pearl}, since our framework disentangles task inference (both global and local) from decision making, the agent in our framework can safely be trained using off-policy RL algorithms. As discussed in Sec. \ref{sec:sac}, we use soft actor-critic algorithm \cite{sac} and further extend the definition of $\pi_\theta$, $\Q_\theta$ and $V_\theta$ to further take the (global and local) latent context variables as input. Following \pearl, we implement the $R$ term in Eq. \ref{eq:pearl} to recover the value function and additionally design a sampler $\mathcal{S}_c$ for context sampling from the replay buffer $\gB$ so that the contexts do not diverge too much from that achieved by the current parameters.
We list the meta-training algorithm of \methodname in Algorithm 1. In order to integrate our joint latent space with SAC, during meta-training time, we first collect trajectories for each task in each iteration as defined in Algorithm 2. Then we sample contexts and batches of trajectories from the replay buffer. We use $\globalenc$ to calculate the global latent context variable and then run $\localenc$ to achieve the local context variable for each step in the trajectory batch before we optimize the loss. During meta-test, we directly roll out our policy while using $\localenc$ to update the local context variable at each step, which is very similar to trajectory collection in Algorithm 2.

\section{RELATED WORK}

Our work is closely related to meta-learning or learning to learn in reinforcement learning settings. 

\textbf{Meta-Learning}.
Meta-learning tries to address the problem of learning to learn \cite{schmidhuber1987evolutionary}. The goal is to leverage the existing knowledge and data to grant the model with more inductive bias so that when facing a set of new tasks, the model can quickly adapt to it \cite{thrun1998learning,bengio1990learning}.  

\textbf{Gradient-based Meta-RL.}
In the context of reinforcement learning, one line of meta-RL models uses gradient-based updates for few-shot adaptation \cite{maml,emaml,promp,xu2018learning}. The objective for meta-training is to find a set of parameters such that they are good initialization for a wide range of tasks. At meta-test time, a few gradient updates can potentially result in high performance, achieving adaptation towards unseen test tasks. The gradient-based meta-RL models do not explicitly infer tasks and hence cannot model the structure of sub-tasks. Another limitation is that this line of work is mostly optimized using on-policy data, thus making the learning process sample inefficient.

\textbf{Meta-RL with Memory.} 
Another line of meta-RL models uses recurrent network structure for the agent \cite{rl2,wang2016learning}. The goal herein is to model previous steps implicitly with the hidden states in the recurrent network. While it is closely related to our work, however, it does not reason over the uncertainty about the task structure and nor does it perform task inference explicitly. This line of work also demonstrates limited sample efficiency due to the on-policy RL update.

\textbf{Context Conditioned Meta-RL.} 
The most related line of work is context based meta-RL models \cite{pearl,iclrpearl,fakoor2019meta}, where the goal is to explicitly perform task inference using contexts.
These models represent tasks with latent context variables and the objective performs context inference as a separate module on top of Q-learning \cite{mnih2015human}. 
The task identification is formalized as a variational inference problem, and they naturally disentangle task inference from decision making by using context conditioned value functions or agents. Integrated with off-policy algorithms, these models achieve significant higher sample efficiency and asymptotic performance in meta-RL. Given a new task, these models make quick adaptation through exploration with posterior sampling \cite{strens2000bayesian}. 
However, in this line of work the latent context variables are fixed across episode and does not allow for compositional tasks with multi-step sub-tasks or sub-goals. 

\textbf{Hierarchical RL.} 
Our work is also related to hierarchical RL \cite{dietterich2000hierarchical}. Hierarchical RL models generally learn policies with hierarchical structure, which often results in a high-level and low-level policy. The high-level policy reasons over the structure of the task and either selects a skill to execute \cite{florensa2017stochastic}, or assigns a goal for the low-level policy \cite{li2019hierarchical,nachum2018data}. Our work focuses on meta-RL and models the task distribution with a global and local context variables rather than directly learn a hierarchical policy \cite{haarnoja2018latent}.

\begin{figure*}[t]
\centering
    \vspace{-5pt}
    \includegraphics[width=\textwidth]{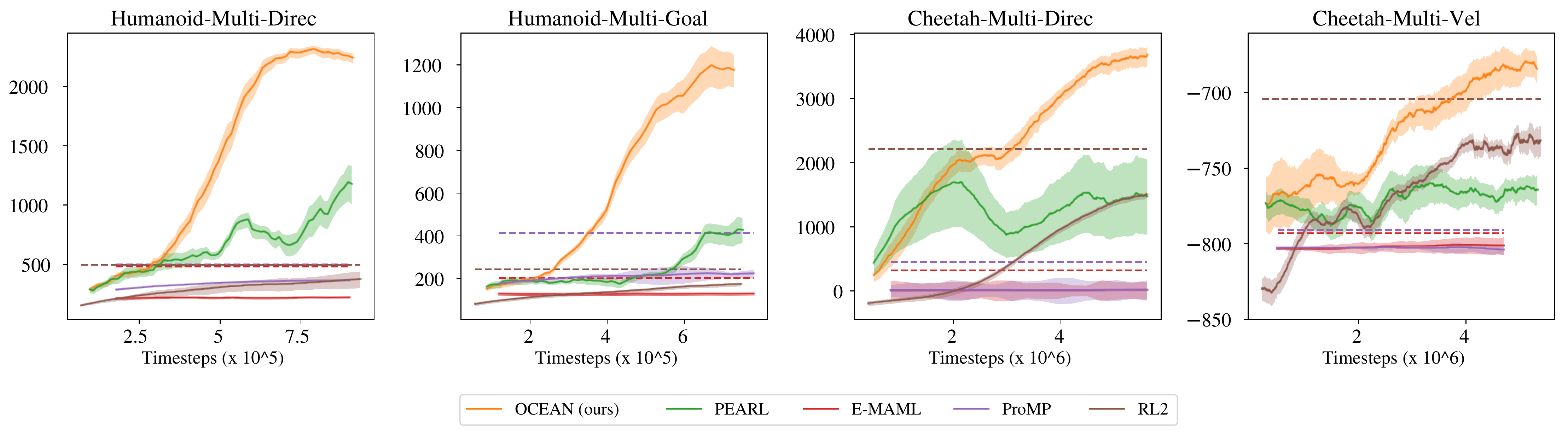}
    \vspace{-20pt}
    \caption{Our framework \methodname v.s. several state-of-the-art baselines in multi-stage tasks. \methodname achieves significantly better sample efficiency and performance since we are able to perform accurate online task inference that fits especially in multi-stage setting where each task requires finishing a sequence of sub-tasks.}
\label{fig:main}
\vspace{-10pt}
\end{figure*}

\methodname is a unified framework for meta-learning in sequential decision making. Our model combines the advantages of both world of recurrent memory and latent contexts. We leverage the contexts to infer the task on a global scale while design a local context encoder with recurrent structure to perform online context adaptation and infer the local sub-task. In addition to this we also enable a richer context modeling with a flexible parameterization of the latent space. These empower the model to learn in a complex setting with structured, compositional tasks. 

\section{EXPERIMENTS}

In this section we aim to investigate the following questions: (1) Can \methodname succeed in several multi-stage meta-RL tasks with the online context adaptation? (2) What's the impact of each module in our framework? (3) Do we need to update the posterior within an episode or can we only update the context variables by directly resampling from the posterior at each step? (4) Do different choices of global latent space matter given prior knowledge of the task distributions? 

\subsection{EXPERIMENTAL SETUP}
We evaluate the performance of our framework \methodname on a 2D point-robot navigation task and five simulated environments in Mujoco \cite{mujoco} with continuous control, four of which have a compositional sub-task structure. We provide the details regarding the tasks as well as the baselines below.

\textbf{Point robot navigation.} The agent in 2D plane aims to navigate to different goals on the edge of a half-circle. 

\textbf{Cheetah-Fwd-Back.} 2D cheetah agent aims to run forward or backward, this environment is not multi-stage and only has 2 tasks.

\textbf{Cheetah-Multi-Vel.} 2D cheetah agent aims to run at goal velocity. One task may contain multiple goal velocities and the steps that the goal velocity shifts are randomly sampled.

\textbf{Cheetah-Multi-Direc / Humanoid-Multi-Direc.} 2D cheetah agent / 3D humanoid agent aims to run in goal direction. One task may contain multiple goal directions and the steps that the goal direction shifts are randomly sampled. 

\textbf{Humanoid-Multi-Goal.} 3D humanoid agent aims to run to several goals. One task may contain multiple goals and the steps that the goal shifts are randomly sampled.

The environments in Mujoco are adapted from previous meta-RL works \cite{maml,pearl}, the main feature is that we added multi-stage sub-tasks in each environment. For all the environments, we adhere to the protocol that we sample a fixed number of training and test tasks before training, after the models are trained on the fixed set of training tasks, we evaluate whether the model is able to quickly adapt to the test tasks.

\textbf{Baselines.}
We compare \methodname with several state-of-the-art meta-RL methods, including gradient-based meta-RL models: E-MAML \cite{emaml}, ProMP \cite{promp}; models with recurrent policy: RL2 \cite{rl2}; and context-based meta-RL model: \pearl \cite{pearl} \footnote{The implementation of the baselines can be found in \url{https://github.com/jonasrothfuss/ProMP} and \url{https://github.com/katerakelly/oyster}.}. 
In all the experiments, we have the same number of latent variables as \pearl. We aim to show that our online task inference scheme can make the best use of the latent variables than using them all to model global contexts.

\subsection{MAIN RESULTS}
\begin{figure}[!t]
\centering
    \includegraphics[width=0.7\columnwidth]{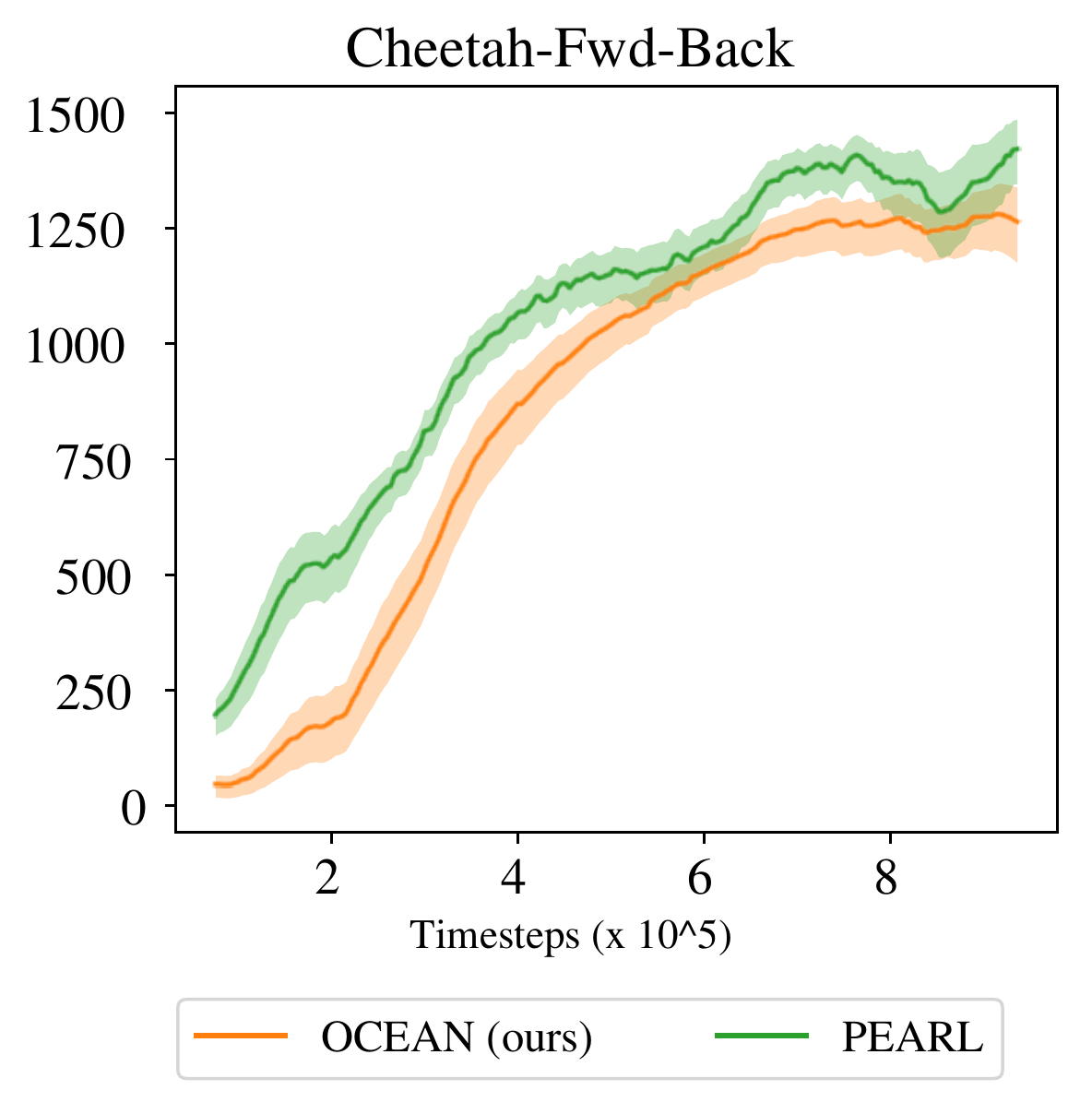}
    \vspace{-10pt}
    \caption{We compare \methodname with \pearl in a task without multi-stage sub-tasks. Since this task does not require modeling the sub-task structure, we observe that \methodname is at par with the task-inference baseline.}
\label{fig:sanity}
\vspace{-10pt}
\end{figure}

We first conduct sanity check on Cheetah-Fwd-Back, where each task is single-stage. We directly compare our model with \pearl. As shown in Fig. \ref{fig:sanity}, our model could achieve comparable results with \pearl although the context variables do not need update in this setting. Our framework essentially makes learning harder as opposed to \pearl. In this setting, \pearl aligns well with the inductive bias that no multi-stage sub-tasks exist. However, our method is much more flexible in the design of latent space and \pearl can be viewed as a special case of our model. If we already have prior knowledge of the task structure, we can reduce our framework to \pearl. 

We further evaluate all the baselines and \methodname in several complex tasks that consist of sequence of sub-tasks, including Cheetah-Multi-Vel, Cheetah-Multi-Direc, Humanoid-Multi-Direc, Humanoid-Multi-Goal. As shown in Fig. \ref{fig:main}, our framework \methodname significantly outperforms all the other baselines in both sample efficiency and asymptotic performance. The final converged performance of the baselines are shown in dashed lines. We observe that RL2 achieves better performance than the gradient-based meta-RL models. The reason is that policy that is based on recurrent model can naturally takes the multi-stage into account, however RL2 cannot reason over the uncertainty of the tasks and also does not have global context variables, both of which make RL2 limited especially when the task structure is complex, for example, in Humanoid-Multi-Direc and Humanoid-Multi-Goal. 
Although in 2D cheetah environments, RL2 is able to achieve comparable and even better results than \pearl, in these 3D environments, RL2 fails miserably due to its incompetence of probabilistic task inference.  

\subsection{ABLATION STUDY}
Here we conduct several ablative experiments to evaluate the importance of each component in our model.

\begin{figure}[!t]
\centering
    \includegraphics[width=0.77\columnwidth]{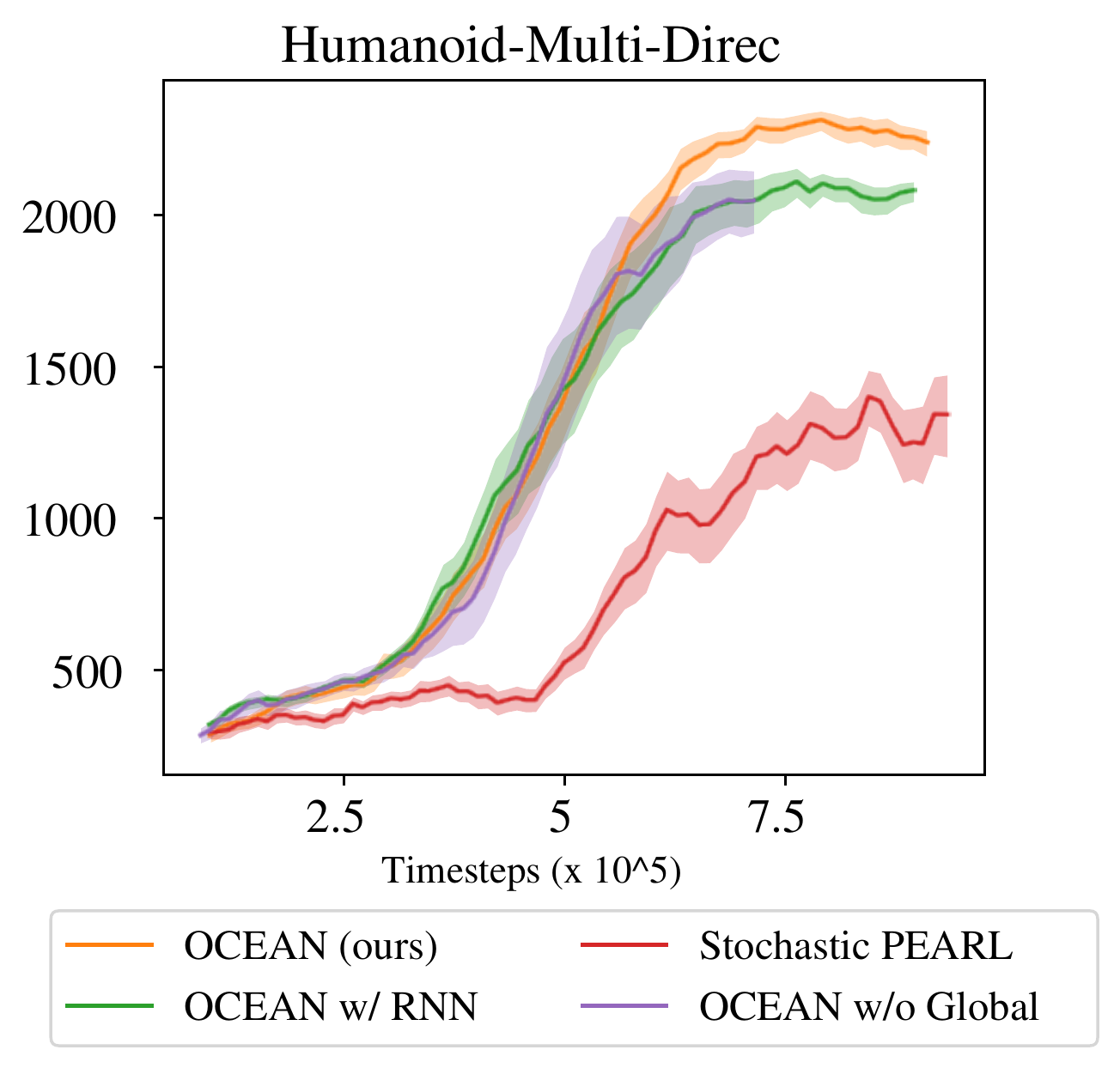}
    \vspace{-5pt}
    \caption{We compare \methodname with several variants on Humanoid-Multi-Direc. }
\label{fig:ablation}
\vspace{-15pt}
\end{figure}

\textbf{Architecture.}
We first ablate the variational recurrent neural network architecture and investigate the benefit of the stochastic transition function as in Eq. \ref{eq:tran} and the dependency between the prior distribution of neighbour steps. We replace the VRNN module with a normal LSTM \cite{lstm} as the architecture of the local context encoder $\localenc$, resulting in \textbf{\methodname w/ RNN}. With a LSTM architecture, the hidden vector $\vh_t$ at timestep $t$ no longer depends on stochastic latent code $\zlocalt$ and the prior distribution of every $\zlocalt$ will be uninformative prior as $\vz_0^{Local}$. As shown in Fig. \ref{fig:ablation}, in Humanoid-Multi-Direc, \methodname w/ RNN can still achieve better performance than baseline \pearl because it is able to adaptively adjust the local context variables, but it is likely to get stuck into local minima and performs worse than \methodname with VRNN architecture as it does not model the dependencies between steps.

\textbf{Global Latent Space.}
Next, we aim to show the importance of global context variables by designing a variant of our framework without global context variables \textbf{\methodname w/o Global}. 
During both meta-training and meta-test, the agent only leverages local context variables to infer the sub-tasks and updates the local context variables online. However, the drawback is that the method cannot leverage the explored contexts (from other episodes) to achieve more information about the task, in other words, the agent does not have a memory buffer and may always start by exploring randomly in the environment and gradually adapt to the task, whereas, our model \methodname can leverage the previous contexts to infer the sub-task structure with a global view, the global context variables also narrow down an agent's exploration area at the beginning of an episode. As shown in Fig. \ref{fig:ablation}, we observe that \methodname w/o Global has higher variance in training than \methodname since it generally requires more exploration steps to gradually take optimal actions.

\textbf{Update Posterior Locally.}
Here we investigate the benefit of adaptively updating the posterior online and design a variant of \pearl with fixed posterior. \pearl infers the posterior of the context variables, samples the context variables from the posterior at the beginning of an episode and holds them constant across the episode. Thus, \pearl can be viewed as freezing the posterior as well as the sample of global context variables. In order to adapt \pearl to the multi-stage setting, one variant of \pearl is to resample the context variables at each step from the same posterior, named \textbf{Stochastic \pearl}. In this sense, the context variables are still updated ``locally''. We list the difference in Table \ref{tab:stochastic}. The result on Humanoid-Multi-Direc is shown in Fig. \ref{fig:ablation}, Stochastic \pearl behaves poorly in this multi-stage task, and barely improves compared with \pearl. The results demonstrate that na\"ively sampling from a fixed posterior does not improve the accuracy of task inference, since it only introduces additional noise instead of effectively reasoning about the compositional structure.

\begin{table}[t]
\resizebox{\columnwidth}{!}{
\begin{tabular}{|c|c|c|c|}
\hline
                 & \pearl & Stochastic \pearl & \methodname  \\ \hline
Posterior        & Fixed & Fixed            & Update \\ \hline
Context Variables & Fixed & Update           & Update \\ \hline
\end{tabular}
}
\caption{Comparison among three methods. \pearl has fixed posterior and context variables; Stochastic \pearl, a variant of \pearl, repeatedly samples the context variables from the fixed posterior at each step; \methodname updates both the posterior and the context variables online.}\label{tab:stochastic}
\vspace{-10pt}
\end{table}

\subsection{INCORPORATE PRIOR KNOWLEDGE OF TASK STRUCTURE}
Here we investigate how to leverage prior knowledge of the task structure by designing the appropriate latent space. We conduct experiments on 2D point robot navigation environments, where the objective of the agent is to navigate to the goal, distributed on the edge of a half circle. In order to create a multi-modal task distribution, we sample goal locations as interpolations between the two end points of the half circle, where the interpolation weight follows Dirichlet distribution $Dir(0.2,0.2)$. Since we already know the goals (tasks) are Dirichlet distributed, we can design a latent space with a Dirichlet prior. Note that since this task is not multi-stage, we did not use the local context variables and only investigate the effect of different global latent space on the performance. For comparison, we also design two alternatives with Gaussian or Categorical distribution as the global latent space. As listed in Table. \ref{tab:robot}, \methodname with Dirichlet prior displays the best performance as it aligns well with the task structure. The categorical distribution on the other hand is not expressive enough for this setting. This result demonstrates that it is beneficial to incorporate the prior knowledge of task structure as inductive bias to the framework for more accurate task inference.

\begin{table}[t]
\centering
\begin{tabular}{|c|c|c|c|}
\hline
Return      & Gaussian & Categorical & Dirichlet  \\ \hline
Point-Robot & 10.0   & 8.5              & 11.7             \\ \hline
\end{tabular}
\caption{The result of \methodname with different choices of global prior distribution in a point-robot navigation task whose goals follow the Dirichlet distribution.}\label{tab:robot}
\end{table}
\section{CONCLUSION AND FUTURE WORK}
In this paper we propose a generalized online task inference framework for meta-reinforcement learning based on online variational inference. The framework consists of global and local latent context variables estimated by two encoders respectively. The global variables capture the general structure of the tasks and can be tailored to the task if the structure is known to the user. The local variables are updated and estimated online across an episode that enables the agent to make smooth transitions from one sub-task to another. Extensive experiments on real-world continuous control tasks have shown superiority of our framework over the state-of-the-art meta-RL methods. One exciting future direction is to train the local context encoders with an auxiliary task to predict the time step of the transition. This auxiliary inference may help the agent in the exploration steps and can also improve the accuracy of sub-task inference. Another future direction is to utilize the contexts from previous episodes in the training of the local context encoder so that we do not need to always start with sampling local context variables from uninformative prior.  

\subsubsection*{Acknowledgements}
{\footnotesize
A.G. is a CIFAR AI chair and also acknowledges Vector Institute for computing support. J. L. is a Chan Zuckerberg Biohub investigator.
We gratefully acknowledge the support of
DARPA under Nos. FA865018C7880 (ASED), N660011924033 (MCS);
ARO under Nos. W911NF-16-1-0342 (MURI), W911NF-16-1-0171 (DURIP);
NSF under Nos. OAC-1835598 (CINES), OAC-1934578 (HDR), CCF-1918940 (Expeditions), IIS-2030477 (RAPID);
Stanford Data Science Initiative, 
Wu Tsai Neurosciences Institute,
Chan Zuckerberg Biohub,
Amazon, Boeing, Chase, Docomo, Hitachi, Huawei, JD.com, NVIDIA, Dell. 
The U.S. Government is authorized to reproduce and distribute reprints for Governmental purposes notwithstanding any copyright notation thereon. Any opinions, findings, and conclusions or recommendations expressed in this material are those of the authors and do not necessarily reflect the views, policies, or endorsements, either expressed or implied, of DARPA, NIH, ARO, or the U.S. Government.
}

\newpage
{\footnotesize
\bibliography{ref}}
\renewcommand{\baselinestretch}{0.95} 
\bibliographystyle{ieeetr}
\newpage

\onecolumn

\appendix

\begin{center}
\begin{huge}
\textbf{Appendix}
\end{huge}
\end{center}

\section{ADDITIONAL ABLATION}
In this section, we perform some additional ablation experiments on a new set of environments to evaluate the performance of \methodname that uses different latent space. Specifically, we tune the complexity of the new environments we proposed. For Cheetah-Multi-Vel ($a$, $b$), we tune the minimum goal velocity $a$ and the maximum goal velocity $b$ in order to control the difficulty of the tasks. For Cheetah-Multi-Direc, this task requires cheetah to run with a sequential backward/forward goal direction, e.g., [Backward, Forward, Backward]. We make the task even more challenging by creating a gap between the meta-train and meta-test tasks. Besides the transition steps, all the combination of goal directions in meta-test tasks are not covered in meta-train. For Humanoid-Multi-Goal-$r$, we tune the radius of the circle where we sample the tasks.

\begin{figure*}[!h]
\centering
    \includegraphics[width=\textwidth]{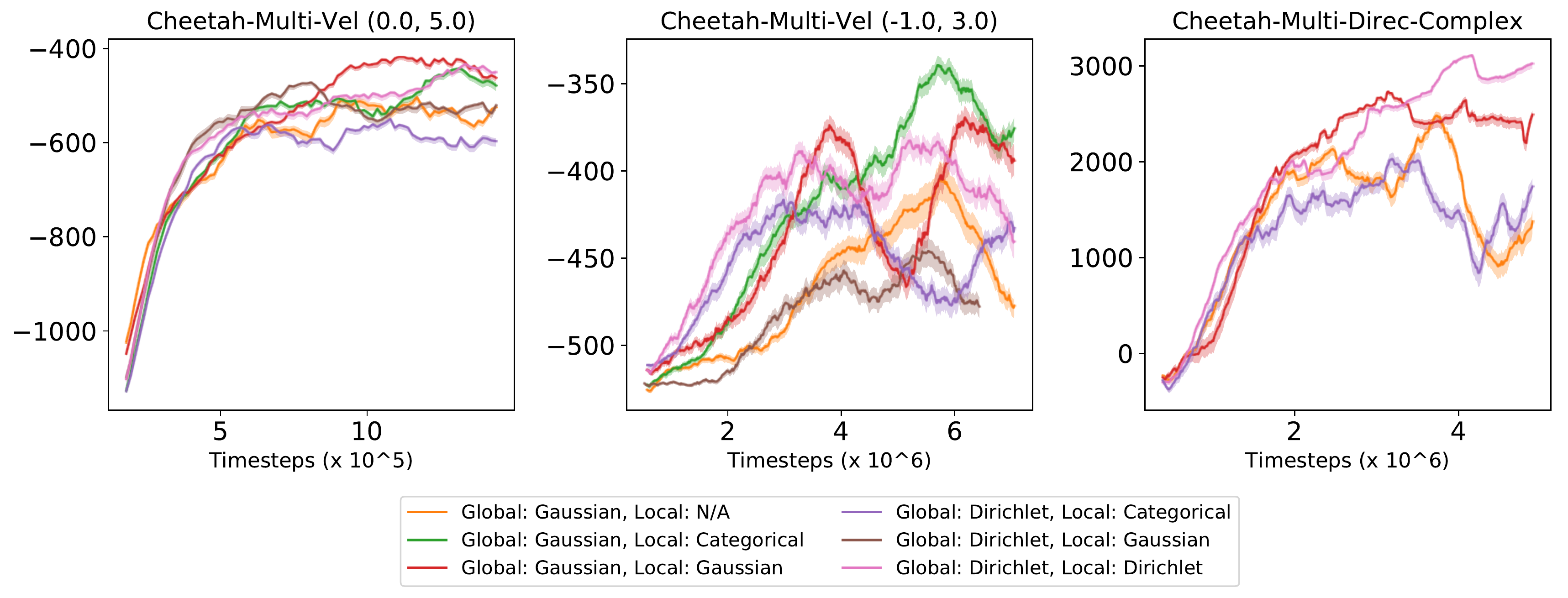}
    \caption{Additional experiments on Cheetah environments.}
\label{fig:cheetah}
\end{figure*}

\begin{figure*}[!h]
\centering
    \includegraphics[width=0.8\textwidth]{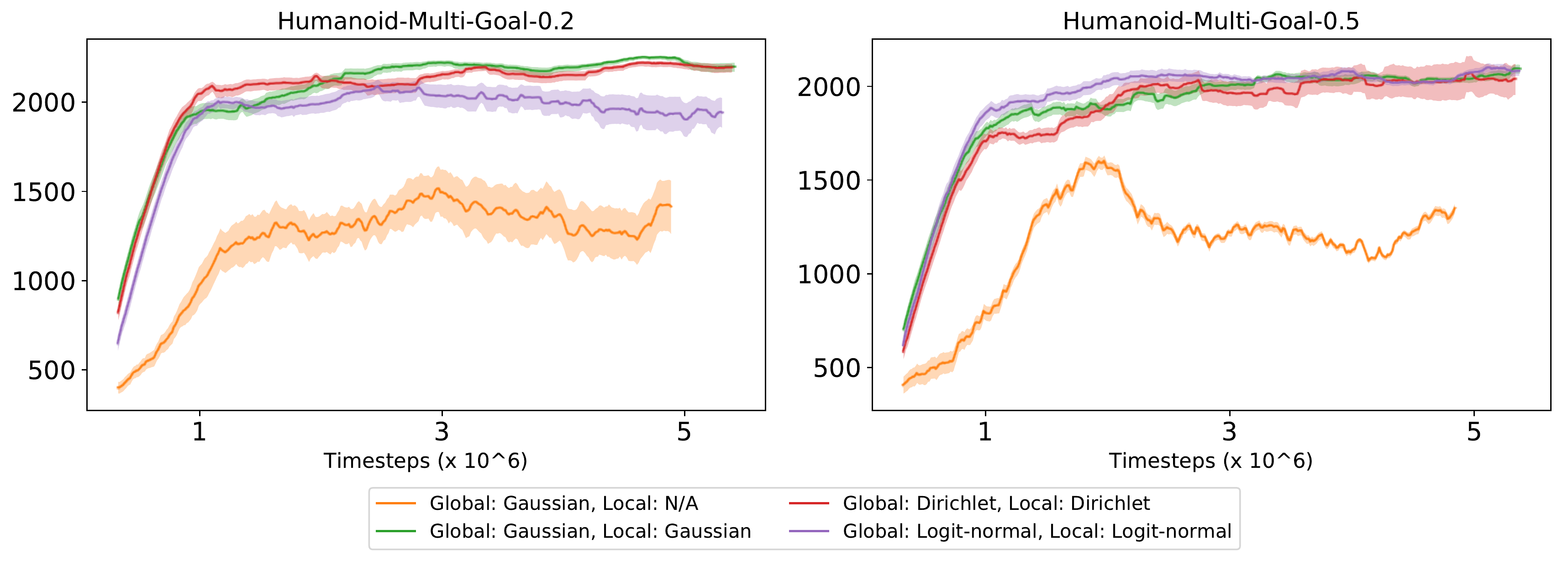}
    \caption{Additional experiments on Humanoid environments.}
\label{fig:humanoid}
\end{figure*}

As shown in Fig. \ref{fig:cheetah} and Fig. \ref{fig:humanoid}, \methodname with local context variables generally works better than \methodname without local context variables (PEARL), since local context variables can better model the transition of the sub-tasks. It often contributes to a boost in both asymptotic performance as well as sample efficiency, especially in Cheetah-Multi-Vel (-1.0, 3.0) and Cheetah-Multi-Direc-Complex, where the transition is clear and the agent cannot simply execute a same policy to approximately finish the complex task. It shows similar trends in Humanoid environments. In Cheetah-Multi-Vel (0.0, 5.0), since all the goal velocities are positive, an agent can learn to go forward with the averaged velocity in order to finish the task. 

\section{EXPERIMENTAL DETAILS}\label{app:detail}
We use Pytorch for \methodname and can be found at \url{https://github.com/pairlab/ocean}. The latent dimension is 12 in total. We use the global latent space to be two Dirichlet distribution, each of dimension 3 to model the mixture of tasks. For local latent space, we use two categorical distribution, each of dimension 3 to model a single on-going sub-task. For fair comparison, the PEARL baseline also has 12 dimension of Gaussian random variables. We train our model for 300 epochs for all the continuous control environment. Each epoch contains 4000 iterations of SAC update. For the point-robot navigation experiment, the latent space we used are 6 dimension of Gaussian distribution versus 3 categorical distribution, each of 2 dimension, versus 3 Dirichlet distribution, each of 2 dimension. Hence, the number of random variables stay the same for fair comparison.

For the continuous control tasks, the number of training and test tasks on Cheetah is 50 and 10; the number of training and test tasks on Humanoid is 15 and 5. Given the small set of training tasks, our task is considered harder than the previous benchmarks, where they do random sampling at the beginning of each episode. The maximum episode length are both 500.

\end{document}